\documentclass[12pt]{article}
 
\usepackage{graphicx}
\usepackage{amsmath}
\makeatletter
\renewcommand*\env@matrix[1][*\c@MaxMatrixCols c]{%
  \hskip -\arraycolsep
  \let\@ifnextchar\new@ifnextchar
  \array{#1}}
\makeatother
\usepackage{amssymb}
\usepackage{setspace}
\usepackage{color}
\usepackage{changes}
\usepackage{tikz}
\usepackage{bm}
\usepackage{cite}
\usepackage{enumerate}
\usepackage{url}
\usepackage{hyperref}

\usepackage{enumitem}
\usepackage{lipsum}
\usepackage{changes}
\usepackage{bm}
\usepackage{changes}
\graphicspath{{figures/}}

\begin{document}

\title{Modular representation and control of floppy networks}

\author{Siheng Chen$^{1}$, Fabio Giardina$^{1}$, Gary P. T. Choi$^{2}$, L. Mahadevan$^{1,3,4\ast}$\\
\footnotesize{$^{1}$School of Engineering and Applied Sciences, Harvard University, Cambridge MA 02138, USA}\\
\footnotesize{$^{2}$Department of Mathematics, Massachusetts Institute of Technology, Cambridge MA 02139, USA}\\
\footnotesize{$^{3}$Department of Physics, Harvard University, Cambridge MA 02138, USA}\\
\footnotesize{$^{4}$Department of Organismic and Evolutionary Biology, Harvard University, Cambridge MA 02138, USA}\\
\footnotesize{$^\ast$To whom correspondence should be addressed; E-mail: lmahadev@g.harvard.edu}
}

\date{}
\maketitle

\begin{abstract} 
Geometric graph models of systems as diverse as proteins, robots, and mechanical structures from DNA assemblies to architected materials point towards a unified way to represent and control them in space and time. While much work has been done in the context of characterizing the behavior of these networks close to critical points associated with bond and rigidity percolation, isostaticity, etc., much less is known about floppy, under-constrained networks that are far more common in nature and technology. Here we combine geometric rigidity and algebraic sparsity to provide a framework for identifying the zero-energy floppy modes via a representation that illuminates the underlying hierarchy and modularity of the network, and thence the control of its nestedness and locality. {Our framework allows us to demonstrate a range of applications of this approach that include robotic reaching tasks with motion primitives, and predicting the linear and nonlinear response of elastic networks based solely on infinitesimal rigidity and sparsity, which we test using physical experiments.} Our approach is thus likely to be of use broadly in dissecting the geometrical properties of floppy networks using algebraic sparsity to optimize their function and performance.
\end{abstract}

Many different physical systems that span multiple scales can be represented as graphs or networks. These include proteins~\cite{jacobs2001protein}, robots~\cite{bernstein1966co,khatib1987unified}, and collections of individual organisms or agents~\cite{camazine2003self,sumpter2010collective,olfati2007consensus}. The physical, chemical and informational properties of these systems such as coordinated motion, communication and control are both enabled and constrained by the nature of the connectivity of the underlying network. Examples include task-relevant representations associated with coordination in motion planning and motor control problems in robotics and neuroscience~\cite{bernstein1966co,khatib1987unified},   spatial segmentation and clustering in biomolecules~\cite{halabi2009protein,he2008hierarchical}, and the mechanics of protein elastic networks~\cite{tirion1996large}. In all these examples, the underlying topological graphs are also automatically endowed with a geometric structure due to the embedding in two or three-dimensional Euclidean space with edges corresponding to the distance between the nodes. This makes the understanding of the spatial structure and organization of the degrees of freedom (DoF) in these networks an issue of critical importance, as learning how to identify, actuate, combine, or remove certain DoF will provide the means to control these systems. 

Since Maxwell's early work in 1864~\cite{maxwell1864calculation}, rigidity theory has been a powerful tool to model these graphs, initially treated as mechanical networks \cite{thorpe1999rigidity}. A key concept there is that of the deviation from isostaticity wherein the network is poised between being rigid and soft because the number of degrees of freedom are just balanced by the number of constraints. Two main approaches are currently used for the study of these networks based on whether they are stiff or compliant. First, for networks with stiff links, the number of DoF can be found using infinitesimal rigidity theory and graph theory~\cite{ jacobs1995generic, guest2006stiffness}, quantities that have been leveraged in network control, metamaterial design~\cite{kim2019conformational, kim2019design, chen2019rigidity, chen2020deterministic}, and determination of the minimal rigid formations~\cite{krick2009stabilisation}. Second, for networks with compliant links, mechanical properties such as bulk and shear modulus can be calculated via molecular dynamics simulation~\cite{broedersz2011criticality}. The intermediate case of networks that transition between stiff and compliant, i.e. can be transformed from mechanisms with floppy modes (zero energy deformations) to structures (finite energy deformations) has been the focus of much recent work from the perspective of phase transitions and critical phenomena. A particular question of interest here is the mechanical response near the jamming transition, or isostaticity (as already defined by Maxwell) and has been extensively studied~\cite{van2009jamming,liu2010jamming,broedersz2011criticality}, with application in allosteric control~\cite{rocks2017designing,yan2017architecture,wodak2019allostery}. The control and tuning process can be done in a bond-by-bond manner, and the bulk and shear modulus can be tuned independently~\cite{goodrich2015principle,hexner2018role}. 
Furthermore, the interaction between geometry, topology, and mechanical properties such as elastic energies has been studied to understand the tangles and structural changes in real-life networks~\cite{liu2021isotopy, dehmamy2018structural}. 

Most of these studies have focused either on over-constrained systems where there are no floppy modes, or systems near the isostaticity transition where few floppy modes are present. Recent work has also demonstrated how to stiffen under-constrained networks by geometric incompatibility~\cite{merkel2019minimal}, or how to actuate a single floppy mode using correlated noise~\cite{woodhouse2018autonomous}. But what if one is far from the isotaticity threshold with a large number of floppy modes? What is the best way to uncover modularity in these floppy modes, as manifest in their spatial structure and represented as some combination of locality, hierarchy, separability, and nestedness, for example? And what is the best way to control these floppy modes? Since actuating all the floppy modes at once is both expensive and inefficient, are there efficient representations that can guide their activation or suppression? Since many natural and artificial networks must start out as under-constrained networks, answers to these questions is likely to be of broad value. Here we present a systematic framework to identify and represent multiple floppy modes, and use this representation to construct protocols to efficiently actuate and control the network. To uncover the hidden hierarchy in the floppy modes, we use the notion of maximum sparsity in the mode representation, defined in terms of the smallest number of non-vanishing {vector} elements in the representation. We then demonstrate the application of these sparse mode representations in a range of different applications: robotic reaching tasks with motion primitives from the sparsity-based representation, and predicting the linear and nonlinear response of elastic networks based solely on infinitesimal rigidity and sparsity.

\section*{Sparse basis representation}
Any Euclidean graph can be represented as a mechanical network with $N$ nodes whose movement is constrained by a set of edges $\mathcal{E}$. If the length of these edges is fixed, this leads to a set of geometric constraints that can be written as 
\begin{equation}
g_i=\|\bm{x}_p-\bm{x}_q\|^2-l_{pq}^2=0,
\end{equation}
where node $q$ is connected to node $p$ by an edge $[p,q] \in \mathcal{E}$, and $\bm{x}_p$ denotes the coordinates of node $p$.
A floppy or zero energy mode is defined as an infinitesimal mode of motion $\bm{dx}$ in the structure which does not violate any constraints. Therefore, for any constraint $g_i$, we have $dg_{i}=\sum_j\partial g_i/\partial x_j~dx_j=0$. By defining the elements of the rigidity matrix $\bm{R}$ by $(\bm{R})_{ij}=\partial g_i/\partial x_j$, all the constraints can be written in terms of a linear matrix relation, 
\begin{equation}
\bm{R}\cdot\bm{dx}=0.
\end{equation}

Therefore, the null space of $\bm{R}$ contains information about all the floppy modes in the system: each  basis in the null space of $\bm{R}$ corresponds to a mode of motion consistent with the  constraints. Each set of linearly independent basis vectors spanning the whole null space provides a possible representation of all the motions, and any non-degenerate rotation matrix can transform one basis to another, so that there are infinitely many equivalent representations of the null space. 

\begin{figure*}[t!]
\centering
\includegraphics[width=\textwidth]{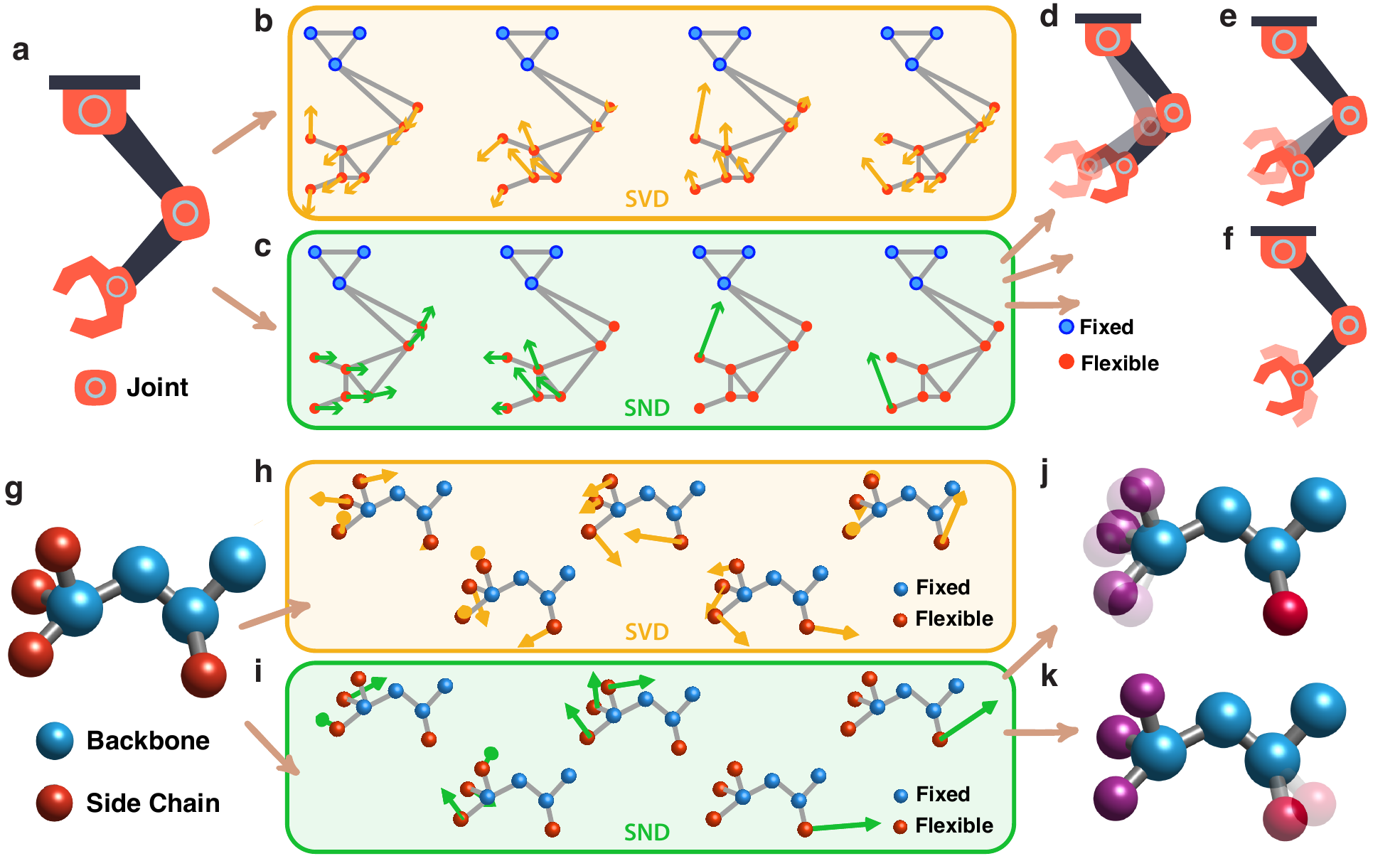}
\caption{\textbf{Mode decomposition in robotics and biology.} \textbf{a}, A simple model robot arm with three joints and four DoF. Abstracted as a node-bond network, the four floppy modes can be represented as in \textbf{b} using the singular value decomposition (SVD) method, or in a hierarchical way as in \textbf{c} using the sparse null space basis decomposition (SND) method. The second method provides a clear separation of all the motions (\textbf{d}, full arm, \textbf{e}, forearm, and \textbf{f}, fingers). \textbf{g}, A biological example of a molecule with backbone atoms (blue, spatially fixed), and side chains (red, flexible). The left three red atoms are pairwise connected (not shown). Using the two methods as in the example above, the five floppy modes are decomposed in \textbf{h} and \textbf{i}. Using the SND method, the motions of the left side chain and the bottom right side chain are clearly separated, as shown in \textbf{j} and \textbf{k}.}
\label{fig:intro}
\end{figure*}

As a simple example, consider a robot arm shown in Fig.~\ref{fig:intro}\textbf{a}, and ignore the movement of the fingers for now. The biologically and technologically natural representation of the modes of motion in this network are the rotation around the base (shoulder), and the rotation around the first joint (elbow) (Fig.~\ref{fig:intro}\textbf{d-e}). However, there are other equivalent representations, e.g. one that consists of (i) allowing the forearm and upper arm to rotate together in the same direction and (ii) allowing them to rotate in opposite directions. On a completely different scale, consider a molecule shown in Fig.~\ref{fig:intro}\textbf{g}. When the backbone atoms are fixed in space, the movement of the side chains on the left and on the bottom right should be independent of each other, resulting in independent rotation modes (Fig.~\ref{fig:intro}\textbf{j-k}). However, it is also possible to represent the motions using a linear combination of the two rotations. In both these examples, we see that while there are mathematically equivalent representations of the null space basis, some representations seem to be intuitively better as they are modular and show spatial structure embodied in locality, nestedness and hierarchy that might make them easier to control.  {We note that the notion of modularity here is different from that in graph theory, defined purely based on links (connections) in the network. A modular representation of floppy modes is a representation where different modes actuate different parts of networks with little overlap or interaction. A modular network can have a representation of floppy modes that is either modular or non-modular depending on the decomposition algorithm, as we shall see later.}  We limit our study to 2D networks without edge crossings and interactions for clarity of presentation and since it is easier to compare to physical experiments with silicone elastic networks. However, our framework can be extended to 3D, since the definition of the rigidity matrix is not restricted to two dimensions.

So how can we find a representation basis that maximizes this modularity and hierarchy? A natural response might be to use the singular value decomposition (SVD) of the rigidity matrix $\bm{R}$ to get a set of zero modes and states of self-stress \cite{pellegrino1993structural, pellegrino1986matrix}. However the zero modes identified from SVD typically involve motion in all coordinates, and are neither modular nor hierarchical (see SI Section~S1.3).

An approach to achieve modularity is to attain maximum sparsity in the zero mode basis, i.e. find those modes whose infinitesimal motions only involve a small number of nodes in a small spatially localized neighborhood. These local modes will also be naturally spatially separated, as any linear combination of them will decrease the resulting sparsity in the bases. The remaining modes, if any, are more global, as they involve motions of more nodes. Thus a classification method based on maximizing sparsity has much potential in helping build a modular hierarchical representation. Mathematically our problem then translates to finding the sparsest null space basis of the rigidity matrix. Finding a sparse null-space basis has been the subject of study in large-scale constrained optimization and is known to be NP-hard~\cite{coleman1986null}, {so that all current attempts to solve this use a variety of heuristic algorithms that take determine which linear recombination derived from the null space of the rigidity matrix is sparse, and hence local, modular and hierarchical.} Here we adapt and generalize a recently proposed algorithm from numerical linear algebra to determine a minimal number of nonzero entries in the null space basis of a matrix  ~\cite{abaffy1989abs} - the Sparse Null-space Basis Decomposition~\cite{khorramizadeh2013efficient}, denoted as SND (see SI Section~S1 for details). We show that it has a natural geometric interpretation in the context of the rigidity matrix, and is perfectly suited to determine the maximally sparse basis for the rigidity matrix $\bm{R}$. Furthermore, we generalize this into a multi-scale method to determine a hierarchy of zero energy modes (in terms of their participation in a motion), this allows us to answer the question of how to determine a sparse, modular and hierarchical representation of the zero energy modes in a graph representation of a range of networks.

\section*{Hierarchical representation and spatial separation of floppy modes} 
To understand how this works in practice, we first consider a simple mechanical network - the robot arm mentioned above (Fig.~\ref{fig:intro}\textbf{a}) that can be abstracted as an under-constrained structure with nodes resembling joints (the shoulder, elbow, and wrist), with the base (shoulder) fixed. Constructing the rigidity matrix from the links in the network, we apply the SND method to find a set of basis {vectors} (floppy modes), and for comparison also use a simple singular value decomposition (SVD) algorithm to generate {another set of} {basis vectors} (see SI Section~S1.3 for a comparison with two other methods). As shown in Fig.~\ref{fig:intro}\textbf{b}, an SVD basis represents a mixed motion: fingers and forearm move together in different directions, and it is hard to interpret the physical meaning of the basis vectors. In contrast, the SND method finds an easily interpretable representation (Fig.~\ref{fig:intro}\textbf{c}) - there are four modes that clearly distinguish between local and global motions. The four modes reflect the movement of the whole arm, the wrist, and the two fingers, respectively, from left to right. Since SND is trying to maximize the sparsity, the simple finger modes, only involving one node (two coordinates), are preferred over more complicated modes.

In addition to separating the modes hierarchically, we now show that the SND algorithm also separates them spatially. Consider the simple molecule (Fig.~\ref{fig:intro}\textbf{g}) with backbone atoms (blue) spatially fixed and side chains (red) being flexible. The three atoms connected to the leftmost blue atoms are connected to each other as a single functional group (these three bonds {between the three red atoms are} not shown in the figure). Since each atom in the side chain has two rotational degrees of freedom, constraint counting shows that there are in total $2\times4-3=5$ DoF. With the SVD method, we can find one representative basis of the five modes shown in Fig.~\ref{fig:intro}\textbf{h}. We see that the modes are hard to interpret, as each of them involves the motion of all the atoms in the side chain. With the SND method, we see another representation shown in Fig.~\ref{fig:intro}\textbf{i}. In contrast to the SVD basis, here we clearly see a separation of the motions of two groups of atoms - the top three modes only involve the functional group on the left, while the bottom two modes only involve the atom in the bottom right. 

\begin{figure*}[!t]
\centering
\includegraphics[width=\textwidth]{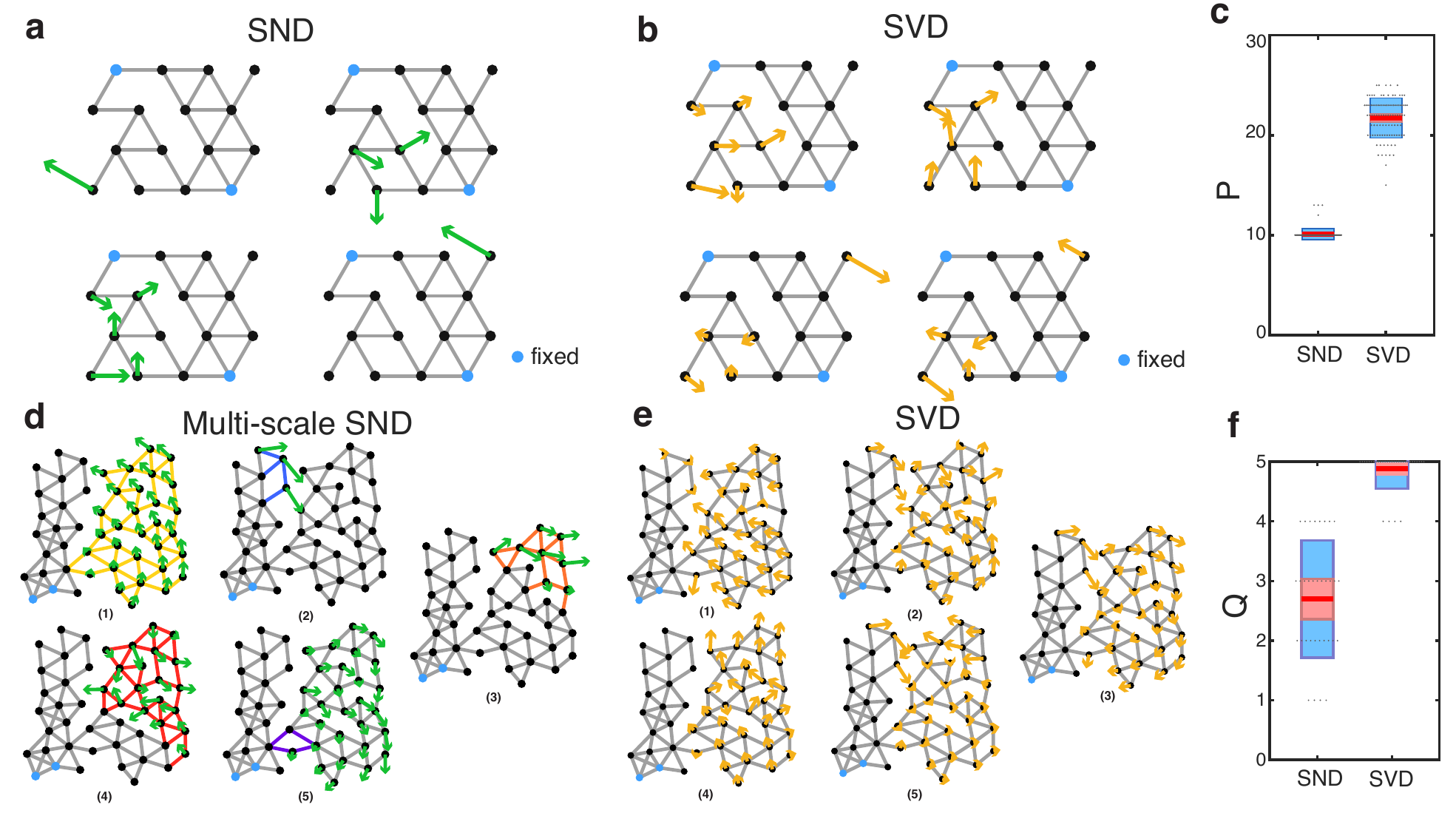}
\caption{\textbf{Mode decomposition and activation in mechanical networks.} \textbf{a}, For this $4\times4$ triangular network, the SND algorithm generates a set of modes with clear hierarchy (from mode~1 to 3) that are spatially separated (mode 4). \textbf{b}, With the SVD algorithm, these modes are not properly separated and it is hard to attach physical meaning to a single mode. \textbf{c}, The participation rate $P$ using the SND method is much lower than that using the SVD method. \textbf{d}, For this complicated network, multi-scale SND can identify the rotational mode (1), and floppy regions (2-5) in a hierarchical manner. \textbf{e} The SVD creates neither hierarchical nor local motions since most nodes are involved in all modes. \textbf{f} The distribution of $Q$ (number of modes a node is involved in) of SND and SVD representation.}
\label{fig:5by5}
\end{figure*}

As just described, there is no scale-dependent hierarchy in deploying the SND algorithm. However, in many real physical systems in protein networks, or robotic systems, there is a need for a scale-dependent approach to determining the optimally sparse and modular representation.  In order to achieve this, we can implement a multi-scale version of the SND algorithm, using the concept of bi-connected components in graph theory to identify all the ``hinges'' in the graph where a section can rotate freely around another. We then apply SND to each sub-graph once these rotational modes are identified. This multi-scale SND ensures that the hierarchy is built-in in the mode representation, and delivers a more intuitive mode representation of the robotic arm shown in Fig. \ref{fig:intro}c, leading to a more intuitive mode involving the rotation of the wrist and the finger such that they rotate together as a rigid section (See SI Section~S2 for more details and an extensive discussion of this multi-scale approach).

To understand the performance of the SND relative to the SVD method, we deploy both on a random planar triangular graph of a $4 \times 4$ network with 4 DoF. The SND method clearly identifies all the floppy modes (Fig.~\ref{fig:5by5}\textbf{a}) in a hierarchical way (the left three modes) and a spatially separated way (the last mode). On the contrary, the SVD method produces a mixture of modes that are hard to interpret (Fig.~\ref{fig:5by5}\textbf{b}). For a quantitative comparison, we define the size $s$ of a mode to be the number of nonzero entries in its vector representation, and the participation rate $P$ of a {set of modes} as the sum of the sizes of all the modes: $P=\sum_j s_j$. By repeating the mode decomposition 100 times (each time with the rows of $\bm{R}$ randomly shuffled), we can calculate the average participation rate $P$ for modes generated by the SND method, and see that it is significantly lower than that of modes generated by SVD (Fig.~\ref{fig:5by5}\textbf{c}). 

{To further demonstrate the intuitive representations arising from SND, we consider a more complex network as shown in Fig.~\ref{fig:5by5}\textbf{d}. The multi-scale SND method can not only identify the rotational mode of the right sub-graph with respect to the left sub-graph (mode (1)), but also identify the floppy regions in a spatially separated (mode (2)) and hierarchical manner (mode (3) to (5)). This representation agrees with our intuition and furthermore reveals the non-trivial orange and red floppy sub-graphs, which are invisible otherwise. In contrast, the modes derived from SVD involve almost all nodes that can possibly move (Fig.~\ref{fig:5by5}\textbf{e}). Defining $Q$ as the number of modes a node is involved in, in Fig.~\ref{fig:5by5}\textbf{f} we see that the representation with SND has a smaller average $Q$ compared to SVD.}

\begin{figure*}[!t]
\centering
\includegraphics[width=\textwidth]{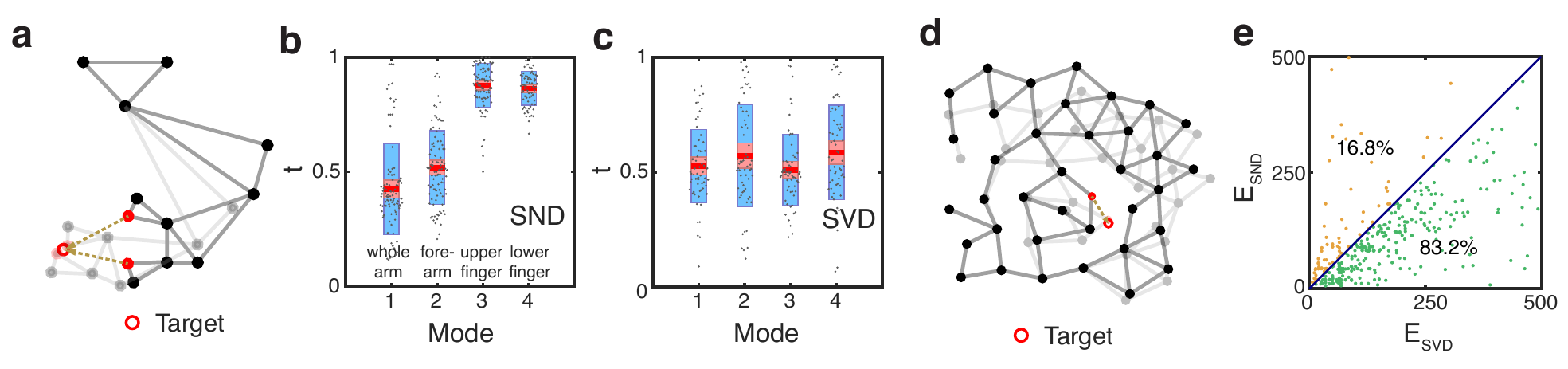}
\caption{\textbf{Mode decomposition and activation in mechanical networks.} \textbf{a} Grasping control task for a simple manipulator arm with two fingers. \textbf{b}, Activation of the sparsity maximizing (SND) modes in the grasping task leads to hierarchical activation of the modes over time from largest (whole arm, mode 1) to intermediate (lower arm, mode 2) to smallest (fingers, modes 3 and 4). Initial arm positions and target locations are randomized. \textbf{c}, Activation of the SVD modes in the grasping task shows no hierarchical activation of modes for randomized initial arm positions and target locations. \textbf{d}, Reaching control task for a random network. \textbf{e}, Energetic cost of the random network control task with motion primitives derived using SVD and SND for randomized target positions and nodes. For the same control task, SND finds a solution that uses less energy 83.2\% of the time. See SI Fig.~S6 for a more detailed comparison.}
\label{fig:reaching}
\end{figure*}

\section*{Reaching control with sparse motion primitives}
The advantages of a sparse mode representation of a floppy network are most naturally seen in the context of network control: Our multiscale generalization of SND has the desirable properties for motion primitives relevant for motor control and robotics. We therefore set out to use these derived modes as motion primitives in a simple grasping task using the robotic arm shown in Fig.~\ref{fig:reaching}\textbf{a} whose task is to pinch a randomly placed target position with its two fingers. The initial arm position is randomized and at every time step, the control algorithm selects and activates the mode which minimizes the distance of the two fingers to the target, the activation time of which is shown in Fig.~\ref{fig:reaching}\textbf{b} and \textbf{c}. Mode 1 corresponds to the activation of the whole arm, mode 2 to the lower arm, and mode 3 and 4 to the two fingers. On average, we see a cascading activation of the modes in time, starting with the largest arm mode and ending with the two finger modes. For comparison, we also use SVD to solve the problem and see no clear distinction can be made in terms of activation time of the modes (see supplementary Movie 3).

The hierarchical activation may appear obvious in the simplified robotic arm, but for complex networks our intuition may fail while modes derived using the sparsest approach can still provide spatially separated and hierarchical motion primitives. We deployed our approach for sparse motion primitives on the random network shown in Fig.~\ref{fig:reaching}\textbf{d} in a reaching control task. A node chosen at random is required to reach a randomized target location. Again we select the mode that minimizes the distance to target at every time step. We track the required  kinetic energy until task completion summed over all nodes for the motion primitives. As shown in Fig.~\ref{fig:reaching}\textbf{e}, for an identical control task, a comparison with the SVD based primitives shows that the use of SND motion primitives leads to a reduced energetic cost to reach the target in $83.2\%$ of the cases compared to the SVD motion primitives. This is perhaps not surprising because most SND modes are local and do not involve the motion of the whole network, and thus minimize irrelevant and redundant motions (see supplementary Movie 4 for an example and SI Fig. S6 for a more detailed comparison). It is worth noting that the SND modes are comparable to motor primitives found using unsupervised learning \cite{todorov2003unsupervised} and show a significant improvement in performance of reinforcement learning agents. Our approach might thus provide a general interpretable  framework for the determination of sparse modes for motion primitives in motor control.

\section*{Control and sequential tuning of a mechanical network}
\begin{figure*}[!t]
\centering
\includegraphics[width=\textwidth]{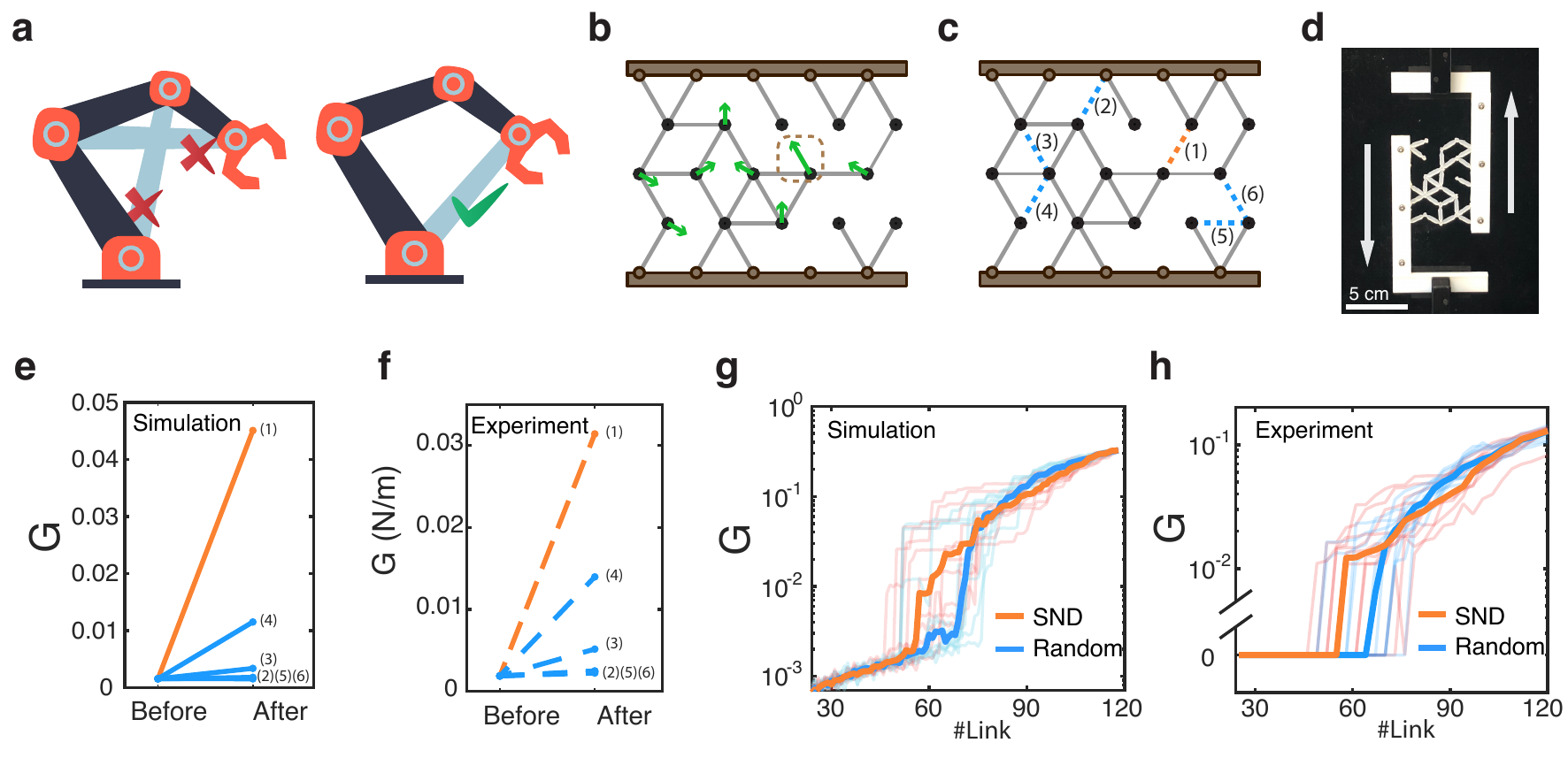}
\caption{\textbf{Efficient rigidification and sequential tuning in mechanical networks.} \textbf{a}, When rigidifying a structure, the most efficient way is to find the ``biggest mode", and freeze it by adding a link to the node with largest movement. \textbf{b}, A triangular network with the largest floppy mode shown in green arrows. The dashed box shows the node with the largest infinitesimal displacement. \textbf{c}, In order to rigidify the network, the orange link (1) can be added. Five other links are randomly chosen ((2) to (6)) for comparison. \textbf{d}, The experimental setup to measure the shear modulus. \textbf{e}, The numerical simulation of the shear modulus of the network (rigid bonds now treated as elastic springs) before and after each of the six links is added. \textbf{f}, The experimental measurement of the shear modulus before and after each of the six links is added. \textbf{g}, For a $7 \times 7$ network, links are added one by one using the efficient (orange) or random (blue) rigidification, respectively. The shear modulus is simulated at each time step for ten runs, and the thick line shows the median. \textbf{h}, The experimental verification of the result in \textbf{g} using a similar set-up as in \textbf{d}.}
\label{fig:triangular}
\end{figure*}

The sparse mode representation not only allows us to control rigid systems, but also provides insight into the control of deformable networks which we turn to next. The separation of the modes into local and global ones provides a simple and efficient guideline for the rigidification of a system. When the network is under-constrained, it is better to first freeze the largest mode (here ``large'' refers to the size of the mode $s$ defined above); within the largest mode, it is better to freeze the node with the largest infinitesimal displacement. A simple robot arm illustrates this idea (Fig.~\ref{fig:triangular}\textbf{a}). To rigidify the whole arm, it is better to freeze the motion of the wrist relative to the base since it has the largest infinitesimal displacement.

{So far, we have based our analysis on the study of the rigidity matrix alone,} and thus have been limited to considerations of infinitesimal motions. In a mechanical setting, this is equivalent to either rigid-bond networks or spring networks with infinite stiffness $k$. We now explore the possibility of generalizing our approach for the control of rigidification of a spring network with finite stiffness $k$. Inspired by the robot arm in {Fig.~\ref{fig:triangular}\textbf{a}}, we propose {identifying} the largest mode (with the largest $s$, i.e. involving the largest number of nodes) and then adding edges {between} the node with the largest infinitesimal displacement and one of its neighbors {to rigidify the network} (which we refer to as the MS protocol, for maximizing stiffness). Intuitively, freezing or rigidifying this large floppy mode adds more low frequency modes to the system, thus making the structure more connected and rigid (See SI Section~S4 for more details, improvements to the protocol, and comparison with other graph theory-based methods.).

To test this idea for efficient rigidification, we first apply it to a $5 \times 5$ network with the link pattern shown in Fig.~\ref{fig:triangular}\textbf{b}, with the biggest floppy mode represented in green arrows, and the node with the largest infinitesimal displacement identified in the dashed box. In order to freeze this mode, link (1) can be added (Fig.~\ref{fig:triangular}\textbf{c}), while links (2) to (6) are randomly chosen among the remaining free links. In order to test whether our MS control protocol from the infinitesimal approach works {in} finitely extensible spring network sense (finite $k$), we use the shear modulus $G$ to characterize the overall rigidity (stiffness). If the spring stiffness is $k$ and the rest length is $l_0$ for all the springs, and the strain is $\gamma$, $G$ can be calculated as follows~\cite{broedersz2011criticality}:
$G=\frac{2}{A}\frac{E}{\gamma^2}, $ where $A$ is the area of the network, and $E$ is the {stretching} energy $
E=\frac{1}{2}\frac{k}{l_0}\sum_{[i,j]\in\mathcal{E}}(\bm{x}_i-\bm{x}_j)^2.$
We also carry out a physical experiment with cast silicone rubber networks (see SI Section~S4.9 for details), with the force measurement setup as shown in Fig.~\ref{fig:triangular}\textbf{d}. In Fig.~\ref{fig:triangular}\textbf{e}-\textbf{f}, we show that both numerical simulations and the experiments confirm that adding link (1) increases the shear modulus much more significantly than adding other links, i.e. the MS protocol based on infinitesimal rigidity remains of value for finitely deformable elastic networks where geometric nonlinearities cannot be neglected. 

With the successful test of the MS protocol at the single bond level, we proceed to apply the MS control protocol on the sequential tuning of system rigidity. As an example, we apply the tuning process to a $7\times7$ triangular network. Starting from 10 network configurations with $20\%$ randomly chosen links, we sequentially add links one by one using the MS protocol and compare the result with a random protocol. With the same number of links added, the network tuned with MS protocol (the orange line) indeed has larger shear modulus than the case where links are added randomly (the blue line) {when the network has a non-zero shear modulus ($50- 70$ links)} (Fig.~\ref{fig:triangular}\textbf{g}, the thick line shows the median for each case). Physical experiments with cast networks (see SI Section~S4.9 for details) verify our numerical results (Fig.~\ref{fig:triangular}\textbf{h}), suggesting our method can be applied to find the links that rapidly stiffen an under-constrained mechanical network. All together, our results show that the infinitesimal approach (on a rigid network with \textit{infinite} $k$) using the rigidity matrix $\bf R$ is valuable even in determining control protocols for a spring network (with \textit{finite} $k$) (See SI Section~S4 for some improvements to the protocol and a more detailed comparison with other methods). 

 {As a side note, }we notice that once the network has some non-zero shear modulus (associated with $80- 110$ links), the random protocol actually outperforms the MS protocol. To understand this, we observe that once a connected percolating cluster has formed, it is better to connect within that cluster to strengthen this network, rather than adding links to freeze the remaining floppy modes, which most likely are near the boundaries. We also notice that when the network has few links (less than $50$), there is a small nonzero shear modulus. This artefact comes from the random noise added to facilitate the simulation (see SI Section S4A for more details).

\section*{Identifying critical links in a mechanical network}

So far, we have shown the power of the hierarchical representation of modes based on infinitesimal rigidity on the control of global network rigidity, as measured in terms of tuning the overall stiffness or shear modulus. Since floppy networks have a large number of zero energy modes, as we tune the global stiffness of the network, the heterogeneity of the local response of the network also begins to vary. We now closely examine the response of individual bonds to external loading and characterize how this heterogeneity might be amplified. In particular, when an under-constrained mechanical system is subject to deformation / external forcing, we ask if it might be possible to detect critical bonds, i.e. bonds subject to more stretching and more prone to failure using our sparse mode representation. 

\begin{figure*}[!t]
\centering
\includegraphics[width=\textwidth]{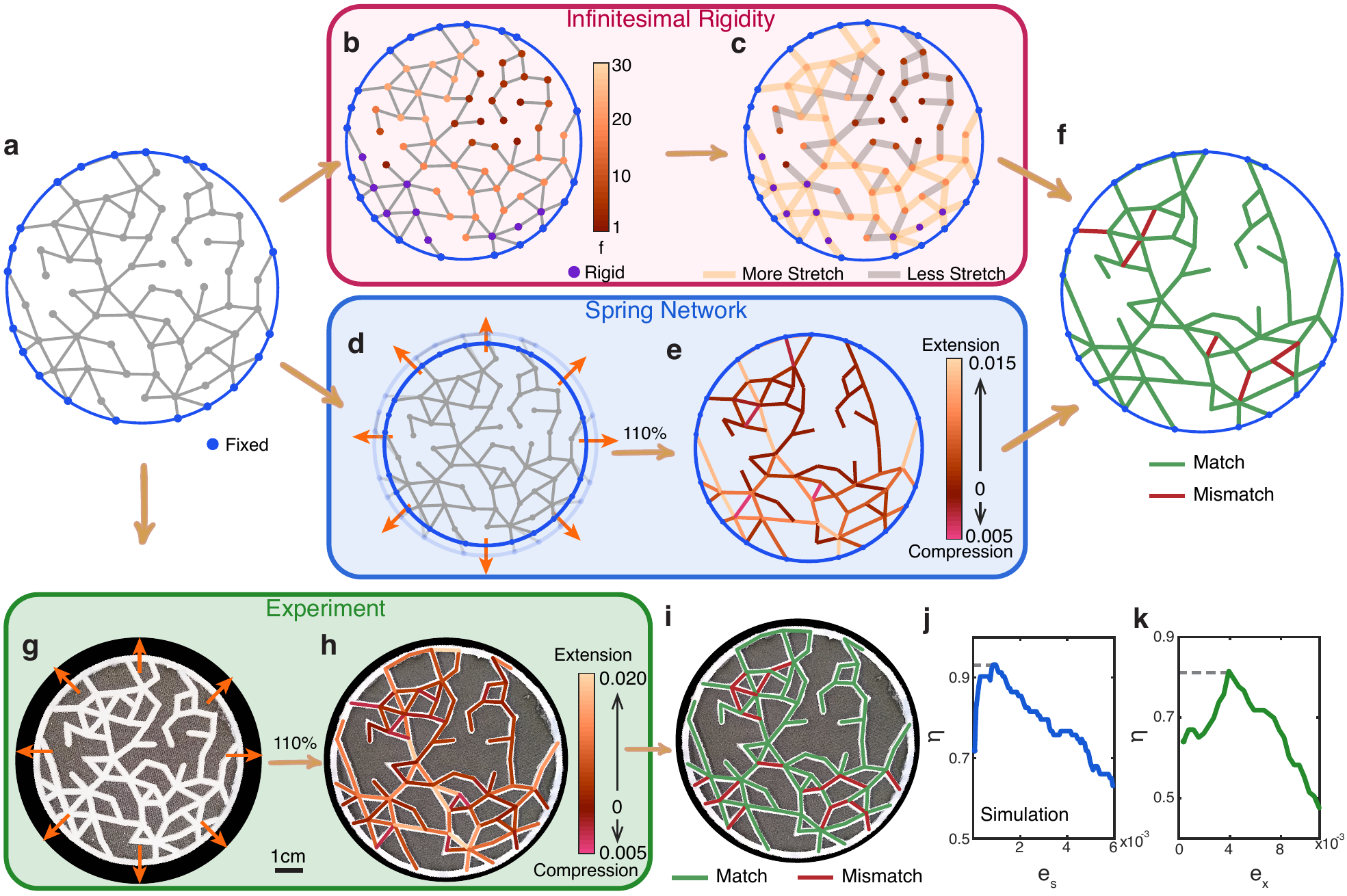}
\caption{\textbf{Load prediction in spring networks.} \textbf{a}, A spring network generated from the standard jamming algorithm. Boundary nodes (blue) are spatially fixed. \textbf{b}, When the network is treated as a frame with rigid bonds, each node is colored based on the globality $f$. The purple nodes are not involved in any infinitesimal modes. \textbf{c}, Prediction of loads: Links connecting nodes that are more ``global'' or rigid nodes are more likely to bear loads (yellow), when the network is deformed. Other links (dark red) are less likely to bear loads. \textbf{d}-\textbf{e}, When the network is treated as a spring network, we apply a uniform outward stretching of 10\% to the boundary nodes (\textbf{d}) and let it equilibrate (\textbf{e}). Brighter links indicate more stretching. \textbf{f}, The prediction from the infinitesimal approach and the numerical simulation agree well. \textbf{g}, Experimental verification: Real network cast with elastic silicone rubber stretched uniformly outwards by 10\%. \textbf{h}, Brighter colors indicate more stretching. \textbf{i}, Similarly, the green color indicate a match between the prediction and the real experiment. \textbf{j}, The matching accuracy between the prediction and the numerical simulation with varying threshold $e_s$. \textbf{k}, The matching ratio between the prediction and the experiment with varying threshold $e_x$.}
\label{fig:circular}
\end{figure*} 

To create a heterogeneous network, we use a bidisperse disk-packing algorithm used to generate jammed configurations~\cite{skoge2006packing,ellenbroek2009non}, and randomly remove some links to allow the structure to have multiple DoF (Fig.~\ref{fig:circular}\textbf{a}). When a uniform outward stretching displacement is applied to this under-constrained network, bonds that can rotate easily do so to accommodate this. For example, in the top right area of the network in Fig.~\ref{fig:circular}\textbf{a} the bonds rotate in response to deformation. To quantify this local flexibility, we use the SND method to determine the sparse representation of the rigidity matrix and find $m$ sets of sparse representations of floppy modes. Defining $s_j^h$ as the size of $j$-th mode in the {$h$}-th {set} (the size of a mode $s_j$ defined previously is the number of nonzero entries in the vector representation) allows us to quantify the globality for a node as 
\begin{equation}
f_i=\frac{1}{m}\sum_{h=1}^{m}\min_{j\in\bm{B}_i^h}s_{j}^h,
\end{equation}
where $\bm{B}_i^h$ is the set of modes that involve the motion of node $i$ in the {$h$}-th {set}. Using this definition, the nodes that are mostly involved in smaller modes (small globality, darker nodes in Fig.~\ref{fig:circular}\textbf{b}) are likely locally flexible, while the nodes that are within larger modes (large globality, brighter nodes in Fig.~\ref{fig:circular}\textbf{b}) are less likely to adjust their positions to accommodate the stretching. There are also nodes that are not involved in any modes (purple nodes in Fig.~\ref{fig:circular}\textbf{b}). These nodes, together with the boundary nodes and global nodes, are likely to be nodes among which connecting links bear more loads (Fig.~\ref{fig:circular}\textbf{c}, links with lighter shades are predicted to bear more load), as these links are less flexible to rotate and adjust. A threshold $t$ in the globality allows us to carry out a binary classification for all the links into ``more stretched'' or ``less stretched'' ones (see Fig.~\ref{fig:circular}c for the classification with $t=12$ and SI Fig. S11 for the effect of different $t$.)

To test our prediction based on infinitesimal rigidity, we numerically simulate the deformation of the network as one made of springs (finite $k$) to determine which links bear loads under uniform stretching (Fig.~\ref{fig:circular}\textbf{d}). After the network equilibrates (Fig.~\ref{fig:circular}\textbf{e}), the difference between the initial and the final length of each spring is calculated (shown via a color gradient {that is equivalent to the force on a spring}). For comparison with the prediction, we set a threshold $e_s$ for the absolute value of the change of length {(scaled by the diameter of the circle)}, and again carry out a binary classification of the links into ``more stretched'' or ``less stretched'' ones. Defining the number of links that are both predicted and simulated to stretch more as $n_b$, the number of links that are both predicted and simulated to stretch less as $n_o$, and the number of total links as $n_t$, we calculate the matching ratio between simulation and prediction as $\eta=(n_b+n_o)/n_t.$ By adjusting the threshold $e_s$, we see that the matching ratio can be as high as {93.2\%} (Fig.~\ref{fig:circular}\textbf{j}, where $t$ is chosen to be 12; see SI Section~S5.5 for more details). To test these results experimentally, we cast an elastic silicone rubber network having the same geometry (Fig.~\ref{fig:circular}\textbf{g}, SI Section~S5.4). By stretching it uniformly outwards by 10\% (Fig.~\ref{fig:circular}\textbf{h}), we can measure how much each link is stretched. Defining the threshold $e_x$ in the spring extension  {(scaled by the diameter of the circle)} above which a link is classified as ``more stretched," we compare the prediction and the experiment (Fig.~\ref{fig:circular}\textbf{i}), and by varying the threshold $e_x$, we find that the maximum prediction accuracy is about 81.6\% (Fig.~\ref{fig:circular}\textbf{k}). {The discrepancy between the simulation and the experiment might come from (1) the experimental networks not being made of perfect linear springs (because of material nonlinearities and elastic energies associated with nodal deformations), and (2) The thickness of the links in the experiment being non negligible, resulting in the errors in length measurement.} While it has recently been shown that purely graph theory based approach can also help {identifying rigid clusters and} predicting the network failure \cite{berthier2019forecasting, berthier2019rigidity}, our approach provides a more general understanding of network flexibility and zero modes. Altogether our approach based on sparse mode representation provides an efficient way to predict local failure in a network, and it does not require consideration of the network energetics.

\section*{Discussion}
Many physical and biological systems can be modeled as under-constrained networks {embedded in real Euclidean space. A necessary step towards functional control of the network minimally} requires a representation of the (topological) connectivity and (geometric) rigidity of the network, as this enables and constrains the efficient transmission of mechanical, electrical, chemical, hydrodynamic information through the network. By couching the quest for hierarchy and modularity in networks in terms of geometric rigidity and algebraic sparsity, {we have framed the question as one of identifying, representing, and understanding the flexibility (floppy modes) in under-constrained networks. By interpreting, adapting and generalizing a sparse-null-space basis algorithm into a multi-scale variant,} we have shown how to create a sparse representation of the floppy modes that uncovers the hidden modularity in the network and describes the combination of hierarchy and spatial localization within the degrees of freedom of the system. Our network approach is agnostic to the application area, and we demonstrate this by using it to determine and deploy minimal motion primitives in robotics, and also use it to predict and control the heterogeneous mechanical response of finitely extensible mechanical networks at the level of both the entire network and individual edges.

It is likely that our methods will find use wherever there is a need for modular representations in under-constrained networks with a given topological and geometrical structure. We chose to work with a Euclidean metric and the associated infinitesimal rigidity matrix given the nature of our geometric/physical networks. More generally, it would be interesting to ask what the results would be if used a linearized constraint for non-Euclidean geometries, or perhaps even stochastic variations thereof. From a practical perspective, how these modular representations might be engineered, learned or optimized in motor control~\cite{bernstein1966co}, motion planning in robots \cite{kirillova2008nma}, design of mechanical networks across scales in DNA assemblies, protein allostery \cite{wodak2019allostery}, and soft materials~\cite{rocks2017designing, yan2017architecture,jacobs2001protein} remain questions for the future. 

\section*{Methods}
Below we describe the methods used for our computational and experimental framework briefly, with further details available in the accompanying Supplementary Information. 

\subsubsection*{Rigidity Matrix}
The number of rows of the rigidity matrix is equal to the total number of constraints. Besides the length constraint of links described in the main text, the spatial confinement is also reflected in the rigidity matrix. For any node $j$ that is spatially fixed,
\begin{equation}
g=\bm{x}_p-\bm{x}_{p_0}=0,
\end{equation}
where $\bm{x}_{p_0}$ is the fixed position for node $p$. 
In the 2D (3D) case, this corresponds to two (three) rows in the rigidity matrix, with only one nonzero entry per row.

All rows are normalized to have unit norm. Random shuffling of the row order is done each time before the SND calculation.

\subsubsection*{The SND algorithm} 
Given a sparse matrix $M \in \mathbb{R}^{m \times n}$, the SND algorithm~\cite{khorramizadeh2013efficient} starts with the identity matrix $H_1 = I_{n \times n} \in \mathbb{R}^{n \times n}$. Then for $i = 1, 2, \dots, m$, the algorithm iteratively updates $H_{i+1}$ such that the rows of $H_{i+1}$ form a basis with a minimal number of nonzeros for the null space of the first $i$ rows of $M$. Ultimately, $H_{m+1}$ becomes a sparse basis for the null space for $M$ (see SI Section~S1 for more details).

\subsubsection*{Spring network simulation}
For the shear modulus calculation, at the beginning of the simulation, the nodes on the top row are displaced to the right by 8\% of the width from top to bottom ($(N_y-1)l_0\times\sqrt{3}/2\times0.08$), and are fixed spatially. The motions of the nodes are simulated 20,000 steps using an overdamped Verlet integration scheme. Random uncorrelated noise (uniformly sampled from $-0.0001$ to $+0.0001$, much smaller than $l_0$) is added to the displacement for each node to facilitate the equilibrium process.

\subsubsection*{Network rigidification}
In the network rigidification process, the MS protocol is as follows: Starting from the initial network configuration,
\begin{enumerate}[label=(\alph*)]
\item Find the set of floppy modes using the SND algorithm.
\item Find the mode with the largest $s$.
\item Within this mode, find the node with the largest infinitesimal displacement.
\item Connect this node to one of its neighbors. If there are multiple closest neighboring nodes available, randomly choose one.
\item (For sequential tuning) Repeat steps (a) to (d) until the network is fully connected.
\end{enumerate}
See SI Section S4 for more variations of the MS protocol.

\subsubsection*{Designing the elastic network}
An initial configuration is generated using an algorithm for random close packing of bi-disperse disks~\cite{skoge2006packing}. Disk centers are transformed into nodes of the network, among which links are mapped from overlapping disks. About 35\% of links are removed thereafter, making the network under-constrained with 18 DoF. We use a circular boundary to avoid having too many free ends. The nodes on the boundary are fixed.

\subsubsection*{Load prediction}
The procedure of the load prediction in Fig.~\ref{fig:circular} is as follows:
\begin{enumerate}[label=(\alph*)]
\item Apply SND to find a set of floppy modes. Calculate the size of each mode $s_j$.
\item For each node $i$, find out the set of modes $\bm{B}_i$ that involve the movement of node $i$.
\item Repeat (a)--(b) $m$ times. Calculate the globality as $f_i=\frac{1}{m}\sum_{h=1}^m\min_{j\in\bm{B}_i^k}s_j^h$. If node $i$ is not involved in any mode, $f_i$ is set to 0.
\item To do a binary classification, set a threshold $t$ for the globality: nodes with $f$ above $t$ are called global nodes. Define the set of eligible nodes as the nodes that are either (1) boundary nodes, (2) fixed nodes, or (3) global nodes. 
\item Start from the boundary nodes and do a breadth-first search (BFS) until reaching another boundary node. The search path where node $i$ and node $j$ are connected has to follow two requirements: (1) node $i$ and node $j$ are connected by a link. (2) node $i$ and node $j$ are both eligible nodes. 
\item Mark all the links along the shortest path as the potential links to bear loads (more stretched). 
\item Repeat (e)-(f) until all the boundary nodes have been visited.
\item All marked links are predicted to bear more stress, and the remaining are predicted to bear less stress.
\end{enumerate}

\subsubsection*{Experiment}
For the network rigidification experiment, physical networks are cast using 3D printed molds and silicone rubber (Dragon Skin\texttrademark ~30, Young's modulus $5.93\times10^5$~Pa). Each network is fully connected initially, and we cut connections sequentially if required by the experimental protocol. Shear forces were measured on an Instron 5566 with a 10N load cell by clamping the outermost edges of the network along its length with a 3D printed clamp and applying a displacement (at a rate of 1~mm/s) to one side until a displacement of 10\% of the network length is reached. More information in SI Section S4.

For the load prediction experiment, a circular network was cast using the same material as above. The physical network diameter is 100~mm and the thickness of an edge is 2.5~mm. The outermost edge of the circular network is glued on a stretchable black spandex cloth which is uniformly stretched to reach an 11\% size increase. Network nodes were extracted and evaluated from image data (see SI Section S5 for more information).

\section*{Acknowledgments}{This work was partially supported by the Swiss National Science Foundation (FG) the National Science Foundation under Grant No.~DMS-2002103 (GPTC), and DMR-2011754, EFRI-1830901, DMR-1922321, the Simons Foundation and the Seydoux Fund (LM).}

\bibliographystyle{ieeetr}
\bibliography{Network-arXiv}

\end{document}